%% file: paper.tex
\documentclass{llncs}
\usepackage{epsf}
\usepackage{graphicx}
\usepackage{url}
\usepackage{xcolor}

\usepackage{multirow}
\begin{document}

\title{Impact of Gender Debiased Word Embeddings in Language Modeling}

\author{Christine Basta and Marta R. Costa-juss\`a}
\institute{Universitat Polit\`ecnica de Catalunya, 08034 Barcelona\\
           \{christine.raouf.saad.basta, marta.ruiz\}@upc.edu}

\maketitle

\begin{abstract}

Gender, race and social biases have recently been detected as evident examples of unfairness in applications of Natural Language Processing. A key path towards fairness is to understand, analyse and interpret our data and algorithms. Recent studies have shown that the human-generated data used in training is an apparent factor of getting biases. In addition, current algorithms have also been proven to amplify biases from data. %Consequently, as word embeddings are trained on such biased data, they may inherit gender stereotypes. 

To further address these concerns, in this paper, we study how an state-of-the-art recurrent neural language model behaves when trained on data, which under-represents females, using pre-trained standard and debiased word embeddings. Results show that language models inherit higher bias when trained on unbalanced data when using pre-trained embeddings, in comparison with using embeddings trained within the task. Moreover, results show that, on the same data, language models inherit lower bias when using debiased pre-trained emdeddings, compared to using standard pre-trained embeddings.  

\end{abstract}

\section{Introduction}

Natural Language Processing techniques and applications have been acquiring increasing attention in the last decade. Artificial Intelligence approaches have recently been highly utilized in such techniques.  Unfortunately, these approaches have proven to have some fairness problems \cite{chiappa:2018}. Bias towards a particular gender, sex orientations or race is a common problem arising in most of machine learning applications. As a consequence, there is an open debate on the topic \cite{leavy:2018}. 

Scientists define bias to happen when a model outcomes differently given pairs of individuals that only differ in a targeted concept, like gender \cite{Lu:2018}.  Therefore, more research is carried out to address this issue. There are approaches, e.g. \cite{tolga:2016}, \cite{vera:2018}, that focus on neutralizing word embeddings (which is the task of learning numerical representation of words) to solve the amplification of bias from data. Other techniques, e.g. \cite{rao:2018},\cite{madaan:2018}, work on debiasing the data at the source before it is consumed in training. Moreover, bias (and gender bias in particular) has been addressed in coreference resolution (which is the task of relating expressions that refer to the same thing), e.g. \cite{zhao:2018} and \cite{rudinger:2018}, showing the effectiveness of measuring and correcting these biases in such tasks.

In this paper, we approach understanding gender bias effect on language modeling. The goal of language modeling is to model the distribution of word sequences. Experiments were carried to understand the influence of data and previously debiased word embeddings on the language model. %Therefore, a fair language model should show the same probability regardless of the stereotypes, e.g. "he is a babysitter" vs "she is a babysitter".

The rest of the paper is organized as follows. Section \ref{sec:background} describes the techniques used to make this paper self-contained, which includes word embeddings, their debiased version and recurrent neural language models. Section \ref{sec:questions} discusses the research questions that we are formulating in this study. Afterwards, section \ref{sec:experiments} illustrates the experimental framework including data and system parameters. Results and conclusions are finally reported in section \ref{sec:results} and \ref{sec:conclusions}, respectively.

\section{Background}
\label{sec:background}

In this section, we are providing a brief overview of the Natural Language Processing techniques used along the paper which are word embeddings and their debiased version and recurrent neural language models.

\subsection{Words Embeddings}
Word embeddings are numerical representations of words and they can be computed by different approaches \cite{mikolov:2013,pennington:2014}. One of the most recent popular approaches is using word2vec \cite{mikolov:2013}. Word2vec has two learning  models: Continuous Bag of Words (CBOW) and Skip-gram. In CBOW, the word representation is predicted given its context, whereas in Skip-gram, the context is predicted given a word. Word2vec first builds a vocabulary from training corpus that is fed into the learning model and thus, learns the vector representations of each word. Word2Vec has an advantage of clustering similar sentences due to the ability to calculate the cosine distance among each word \cite{ma:2015}. 

\subsection{Debiased Word Embeddings}
While word embedding models have become essential in Natural Language Processing applications, they have shown some shortcomings.  
Given that word embeddings are learned from human-generated  corpora, they have been shown to learn and, even worse, amplify social biases \cite{tolga:2016,caliskan:2017}.

 Authors in \cite{tolga:2016} were the first to pay attention to the fact that the embeddings themselves have a kind of bias that should be resolved.  To reduce the bias in an embedding, the embeddings of gender neutral words were changed, by removing their gender associations. A post processing method is applied that projects gender-neutral words to a subspace which is perpendicular to the gender dimension, defined by a set of gender definition words \cite{zhao:2018}.
%\item Gn-Glove: GN-GloVe \cite{caliskan:2017} represents protected  attributes  in  certain  dimensions while neutralizing the others during training. As the information of the protected attribute is restricted in certain dimensions, it can be removedfrom the embedding easily. By jointly identifying gender-neutral words while learning word vectors,

\subsection{Recurrent Language Model}
Language model is the task of computing the probability of sequences of words. One popular strategy of computing language modeling has been by using ngram models \cite{stolcke:2002}, which is mainly based on computing statistics of the appearances of chunks of words. An alternative methodology has been using feed-forward \cite{bengio:2003} or recurrent neural networks \cite{mikolov:2010}. The main advantage of the neural networks approaches is employing word embeddings, which means their ability to use classes of words instead of using the words themselves. While the significant advantage of the recurrent networks is taking into account an unlimited number of words, whereas classical ngram models or feed forward networks have to use Markov assumptions limiting the context of a word to a size of a pre-defined window. In practice, vanilla recurrent neural networks are not enough to keep the information of long sequences, that is why gated methods such as long-short term memories (LSTMs) \cite{lstm} or gated recurrent units (GRUs) \cite{Cho:2014} have been recently preferred.

\section{Research questions}
\label{sec:questions}

\begin{figure}
    \centering
    \includegraphics[width=0.75\textwidth]{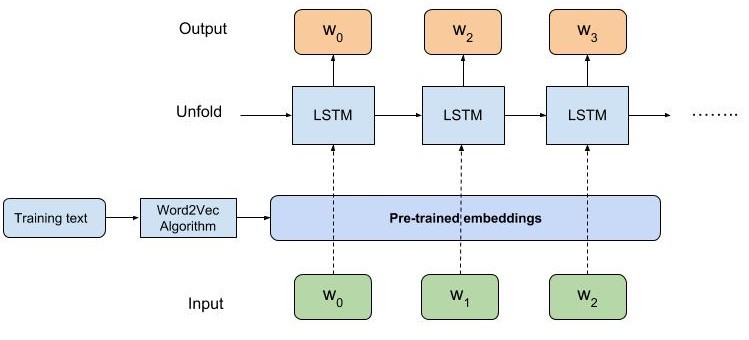}
    \caption{LSTM language model using pre-trained embeddings}
    \label{fig:lstmmodel}
\end{figure}

Language modeling is trained on a large monolingual corpora. State-of-the-art neural language models either learn word embeddings inside the model or use pre-trained word embeddings \cite{makarenkov:2016}. Figure \ref{fig:lstmmodel} outlines LSTM language modeling when using pre-trained embeddings as input to the model.

In this work, we approach answering the following questions:

\begin{enumerate}
\item Do standard language modeling training sets contain bias?

\item Does language modeling show gender bias if it is trained on a biased corpora?

\item Do current debiasing word embedding techniques reduce gender bias in language modeling?
\end{enumerate}

To investigate these questions, we propose to train a recurrent language model and contrast its performance on different test sets with and without stereotypes. Next section describes our experiments.

\section{Experimental Framework}
\label{sec:experiments}

This section reports details on the experimental framework of our study. We report details on the data and implementation used.

\subsection{Datasets}

One standard training set for language modeling is Wikitext-103, in which text is extracted from Wikipedia, consisting of 103,227,021 tokens and 267,735 vocabulary size \cite{merity:2016}. 

\begin{figure}
    \centering
    \includegraphics[width=0.5\textwidth]{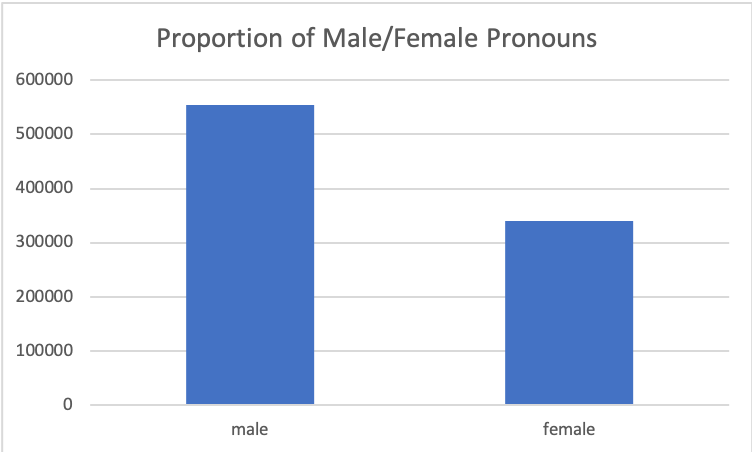}
    \caption{Proportion of Male/Female Pronouns in the training data}
    \label{fig:malefemale}
\end{figure}

To answer the first research question of this study (\textit{Do standard language modeling training sets
contain bias?}), pronouns counting is computed and shown in Figure \ref{fig:malefemale}. The counting shows a clear under-representation of female pronouns. Female pronouns include \textit{she, her, herself}; and male pronouns include \textit{he, his, himself}.

For testing, we used the stereotypes present in occupations. We composed 6 different test sets taking information from \cite{tolga:2016} and which are defined in Table \ref{tab:testsets}.

\begin{table*}[h!]
\begin{center}
\begin{tabular}{l|l|l}
\bf Test  & \bf Structure & \bf Examples  \\ \hline
1 & Definitional male & he is a father, he is a male \\ 
2 & Definitional female & she is a mother, she is a female\\
3 & Stereotypical nouns for male & he is a surgeon, he is a ballplayer \\
4 & Swapping the male's stereotypes & \multirow{2}{*}{she is a surgeon, she is a ball player}\\ & with female pronouns &\\
5 & Stereotypical nouns for female & she is a hairdresser, she is a ballerina \\
6 & Swapping the female's stereotypes & \multirow{2}{*}{he is a hairdresser, he is a ballerina}\\ & with male pronouns & \\
\hline
\end{tabular}
\end{center}
\caption{\label{tab:testsets} Test sets composition }
\end{table*}

\begin{table*}[h!]
\begin{center}
\begin{tabular}{l|l|l|l|l}
\bf Test Set & \bf  Sentences & \bf  Words & \bf  Vocab & \bf  OOV   \\ \hline
1, 2 & 37 & 150 & 42 & 4 \\ 
3, 4 &  207 & 828 & 212 & 21 \\
5, 6 &  60 & 240 & 65 & 5 \\ 
GAP & 2000 & 143155 & 33221 & 15917\\\hline
\end{tabular}
\end{center}
\caption{\label{tab:statistics} Test sets statistics }
\end{table*}

For further evaluation, the models were experimented on a recent balanced dataset: the GAP-Coreference test-set \cite{gap}. This set is available from Google on github \footnote{https://github.com/google-research-datasets/gap-coreference}. As reported, GAP is a gender-balanced dataset containing 8,908 coreference-labeled pairs. We extracted the text from the uploaded raw text to be suitable for our evaluations. Statistics of this set are shown in Table \ref{tab:statistics}, given number of sentences, words, vocabulary (Vocab), and out-of-vocabulary (OOV).

\subsection{Parameters}

Regarding the words embeddings and the corresponding debiasing, we used the Gensim library \cite{rehurek_lrec} for training word2vec CBOW algorithm \cite{mikolov:2013} on the training corpus to create the related words embeddings. We used the implementation from Bolukbasi et al.\cite{tolga:2016} available from github\footnote{https://github.com/tolga-b/debiaswe} to debias these embeddings.

We used the language model implementation from Socher et al. \cite{merityRegOpt,merityAnalysis} available from github\footnote{https://github.com/salesforce/awd-lstm-lm}. Parameters are reported in Table \ref{tab:lmparam}.

\begin{table}[ht]
\begin{center}
\begin{tabular}{l|l}
\bf Parameters  & \bf Value \\ \hline
Batch size & 100 \\
Embedding size & 267734 \\
Embeddings' dimension & 400 \\
Dropping-out rate & 0.2 \\
Sequence length & 70 \\
\hline
\end{tabular}
\end{center}
\caption{\label{tab:lmparam} Parameters in Language model. }
\end{table}
  
\section{Results}
\label{sec:results}

Since we are using the same vocabulary for all our experiments, we relied on the perplexity (PP) which is an standard measure to evaluate the quality of the language model. Table \ref{tab:ppl} shows the results over the different test sets presented in previous section. This table allows us to answer the two remaining research questions, previously mentioned. Regarding the question: \textit{Does language modeling show gender bias if it is trained on a biased corpora?}, the answer is affirmative as shown by the increase in perplexity from definitional \textit{she} (test 2) compared to definitional \textit{he} (test 1). Regarding the question: \textit{Do current debiasing word embeddings reduce gender bias, compared to using pre-trained biased word embeddings?}, the answer seems to be affirmative again. This is shown by the lower relative increment in perplexity from definitional \textit{he} to definitional \textit{she} and the lower difference in perplexity when using the counterpart of either feminine, or masculine stereotypes -i.e. differences between test 1 and 2, test 3 and 4, and test 5 and 6, respectively. It is worth mentioning that the lowest relative increment occurs when using non-pretrained embeddings at all, which corresponds also to the best performing model by far.

\begin{table*}
\begin{center}
\begin{tabular}{lllllll}
System & Test 1 & Test 2 & Test 3 & Test 4 & Test 5 & Test 6 \\ \hline
Non Pre-train Emb & 204.7 & 238.8 (+0,17) & 247.3 & 307.6 (+0,24) & 311.8 & 282.6 (-0,10)\\ \hline
Biased Pre-train Emb & 345.7 & 524.6 (+0,52) & 402.3 & 515.1 (+0,28) &598 & 469.8 (-0,22)\\ 
DeBiased Pre-train Emb & 331.9 & 499.1 (+0,50) & 377.8 & 481.3 (+0,27) & 566.5 & 447.2 (-0,21)\\ \hline
\end{tabular}
\end{center}
\caption{\label{tab:ppl} Perplexity on different test sets}
\end{table*}
%For test 4 (swapping the male’s stereotypes %with female pronouns) and test 6
%(swapping the female’s stereotypes with male pronouns), we see an impact in lan-
%guage modeling when using debiased word embeddings. Examples of this impact
%are shown in Table \ref{tab:examples} which shows sentences and the corresponding perplexity
%for each sentence. The perplexity for these examples gets lower when using the
%debiased word embeddings and this reduction in perplexity can be interpreted
%as a neutralization of stereotypes. The reduction in perplexity in test 4 is higher
%than that of test 6, which means that the effect of debiasing embeddings has a
%higher effect on females than on males. This maybe due to the fact that females
%are underepresented in the training corpus as shown in Figure \ref{fig:malefemale}, making the debias in word embeddings more relevant in this case.
For further analysis of test 4 (swapping the male's stereotypes with female pronouns) and test 6 (swapping the female's stereotypes with male pronouns),  we show examples in Table \ref{tab:examples}. Table \ref{tab:examples} (top) shows sentences from both test 3 and 4 with the corresponding perplexities before and after debiasing embeddings, while Table \ref{tab:examples} (bottom) shows sentences from test 5 and 6. The perplexity for these sentences gets lower when using the pre-trained biased word embeddings and this reduction in perplexity can be interpreted as a neutralization of stereotypes. In both tables, the reduction of perplexity, in test 4 and 6, is higher than the reduction in test 3 and 5, respectively. 
However, the reduction in perplexity in test 4 is higher than that of test 6, affecting more stereotypes, emphasizing the conclusion that debiasing embeddings has a higher effect on females than on males. This correlates to the fact that females are underrepresented in the training corpus as shown in Figure \ref{fig:malefemale}, making the debias in word embeddings more relevant in this case.

\begin{table*}[ht]
\begin{center}
\begin{tabular}{l|l|l|l|l|l}
\bf Sentence (M)  & \bf Bias & \bf DeBias & \bf  Sentence (F)  & \bf Bias & \bf DeBias \\ \hline 
he is an archaeologist & 740.0 & 687.7 & she is an archaeologist & 937.9 & 878.6 \\
he is a ballplayer & 1094.7 & 973.3 & she is a ballplayer & 1328.3 & 1253.6 \\
he is a broadcaster & 467.8 & 428.2 & she is a broadcaster & 593.7 & 542.2 \\
he is a cardiologist & 949.6 & 883.6 & she is a cardiologist & 1222.5 & 1120.6 \\
he is a custodian & 578.5 & 543.3 & she is a custodian & 753.8 & 700.1  \\ 
he is an economist & 716.8 & 662.5 & she is an economist & 926.5 & 865.9 \\
he is a lawmaker & 818 & 747.3 & she is a lawmaker & 1057.4 & 962.1 \\ 
he is a parishioner & 930.9 & 855.2 & she is a parishioner & 1174.5 & 1068.4 \\ 
he is a photojournalist & 914.5 & 829.2 & she is a photojournalist & 1224.5 & 1127.2 \\ 
he is a protege & 534 & 493.4 & she is a protege & 684.3 & 622.5 \\
he is a provost & 603.2 & 562.9 & she is a provost & 787.4 & 731 \\
\hline \hline \bf Sentence (F)  & \bf Bias & \bf DeBias & \bf  Sentence (M)  & \bf Bias & \bf DeBias \\ \hline 
she is a dermatologist & 1161.8 & 1094.7 &
he is a dermatologist & 921.9 & 887.7 \\
she is an organist & 907.6 & 861.2
& he is an organist & 743.8 & 693.6\\
she is a paralegal & 1181.2 & 1096.4
& he is a paralegal & 912.6 & 839.7 \\
she is an observer & 786.5 &745.1 &
he is an observer & 624.9 & 581.6\\
\hline
\end{tabular}
\end{center}
\caption{\label{tab:examples} Perplexity variation in male and female stereotyped sentences from biased to debiased embedding }
\end{table*}

%\begin{table*}[ht]https://www.overleaf.com/project/5c1a6f653331c10b0058012a
%\begin{center}
%\begin{tabular}{|l|l|l|l|l|l|}
%\hline \bf Sentence (F)  & \bf Bias & \bf DeBias & \bf  Sentence (M)  & \bf Bias & \bf DeBias \\ \hline 
%she is a dermatologist & 1161.8 & 1094.7 &
%he is a dermatologist & 921.9 & 887.7 \\
%she is an organist & 907.6 & 861.2
%& he is an organist & 743.8 & 693.6\\
%she is a paralegal & 1181.2 & 1096.4
%& he is a paralegal & 912.6 & 839.7 \\
%she is an observer & 786.5 &745.1 &
%he is an observer & 624.9 & 581.6\\
%\hline
%\end{tabular}
%\end{center}
%\caption{\label{tab:examples2} Perplexity variation in female stereotyped sentences from biased to debiased embedding }
%\end{table*}

Finally, evaluation of the models on the GAP test-set is shown in Table \ref{tab:ppl_balanced}. Perplexity results show the biased and debiased word2vec embeddings perform similarly on a balanced data.

\begin{table*}
\begin{center}
\begin{tabular}{l|l}
System & Perplexity \\ \hline
Non Pre-train Emb & 340.51 \\ \hline
Biased Pre-train Emb  & 1343.7\\ 
DeBiased Pre-train Emb & 1343.6\\ \hline
\end{tabular}
\end{center}
\caption{\label{tab:ppl_balanced} Perplexity on a balanced test set}
\end{table*}

\section{Conclusions and further work}
\label{sec:conclusions}
Generally, human-generated data under-represents females. Debiased pre-trained word embeddings have a decreasing effect on gender bias, when compared to the corresponding biased pre-trained embeddings. However, using self-trained word embeddings in language modeling results in the less biased and best-performing system. 

Further directions in research include studying effect of debiasing embeddings on other Natural Language Processing applications. Moreoever, studying ways to balance data to equally represent males and females. Another approach to be considered, is to investigate how to debias Natural Language Processing techniques. First, these techniques should discover if there is any type of bias, then should start balancing this bias within the system itself.

\section*{Acknowledgments}

This work is supported in part by the AGAUR through the FI PhD Scholarship; the Spanish Ministerio de Econom\'ia y Competitividad, the European Regional  Development  Fund  and  the  Agencia  Estatal  de  Investigaci\'on,  through  the  postdoctoral  senior grant Ram\'on y Cajal, the contract TEC2015-69266-P (MINECO/FEDER,EU) and the contract PCIN-2017-079 (AEI/MINECO).

\bibliographystyle{splncs}
\bibliography{paper}
\end{document}